\documentclass{bmvc2k}
\usepackage{multirow}
\usepackage{here}
\usepackage{graphicx}
\usepackage{tikz}
\usepackage{subfig}
\usepackage{wrapfig}

\usetikzlibrary{positioning, shapes.geometric, arrows.meta}

\title{PseudoCal: Towards Initialisation-Free Deep Learning-Based Camera-LiDAR Self-Calibration}

\addauthor{Mathieu Cocheteux}{mathieu.cocheteux@hds.utc.fr}{1}
\addauthor{Julien Moreau}{julien.moreau@hds.utc.fr}{1}
\addauthor{Franck Davoine}{franck.davoine@hds.utc.fr}{1}

\addinstitution{
\mbox{Université de technologie de Compiègne}\\
CNRS, Heudiasyc Laboratory\\
Compiègne, France
}

\runninghead{Cocheteux \emph{\etal}}{INITIALISATION-FREE C2L SELF-CALIBRATION}

\def\etal{\emph{et al}\bmvaOneDot}

\begin{document}

\maketitle

\begin{abstract}
Camera-LiDAR extrinsic calibration is a critical task for multi-sensor fusion in autonomous systems, such as self-driving vehicles and mobile robots. Traditional techniques often require manual intervention or specific environments, making them labour-intensive and error-prone. Existing deep learning-based self-calibration methods focus on small realignments and still rely on initial estimates, limiting their practicality. In this paper, we present PseudoCal, a novel self-calibration method that overcomes these limitations by leveraging the pseudo-LiDAR concept and working directly in the 3D space instead of limiting itself to the camera field of view. In typical autonomous vehicle and robotics contexts and conventions, PseudoCal is able to perform one-shot calibration quasi-independently of initial parameter estimates, addressing extreme cases that remain unsolved by existing approaches.
\end{abstract}

\section{Introduction}
\label{sec:intro}

Camera to LiDAR extrinsic calibration is crucial for enabling seamless sensor fusion and comprehensive environmental understanding in autonomous systems such as self-driving vehicles and mobile robots. The objective of this task is to determine the 6D transformation \textit{T} between the coordinate systems of a camera and a LiDAR (that is, rotation and translation). Calibration techniques based on traditional vision methods achieved accurate results~\cite{bileschiFullyAutomaticCalibration2009,zhangExtrinsicCalibrationCamera2004a,rodriguezf.ExtrinsicCalibrationMultilayer2008,yuanPixellevelExtrinsicSelf2021,beltranAutomaticExtrinsicCalibration2022,ishikawaLiDARCameraCalibration2018,geigerAutomaticCameraRange2012, pandeyAutomaticTargetlessExtrinsic}, but often require labour-intensive manual procedures, specific environments, or targets. This leads to potential inaccuracies and inefficiencies. Although deep learning-based calibration methods~\cite{schneiderRegNetMultimodalSensor2017,iyerCalibNetGeometricallySupervised2018,jingDXQNetDifferentiableLiDARCamera2022,liuSemAlignAnnotationFreeCameraLiDAR2021,lvLCCNetLiDARCamera2021,wuThisWaySensors2021,cocheteuxUniCalSingleBranchTransformerBased2023} have emerged as powerful alternatives, they are limited by their dependence on initial parameter knowledge (approximation of \textit{T}).

To address these challenges, we propose PseudoCal, a novel sensor calibration method that capitalises on the pseudo-LiDAR concept~\cite{wangPseudoLiDARVisualDepth2020}. It enables accurate and efficient calibration independent of initial parameter knowledge. Existing deep learning-based methods rely on a LiDAR projection in the camera image, based on an initial approximate knowledge of the extrinsic calibration parameters. Thus, they depend on the availability and accuracy of this initial knowledge, and discard most of the LiDAR point cloud, which is not projected into the camera field of view. By leveraging the pseudo-LiDAR concept into the calibration process, PseudoCal is able to work directly in 3D space, dismissing the reliance on initial parameter knowledge.

\begin{figure}[t]
\begin{center}
\includegraphics[width=0.98\textwidth]{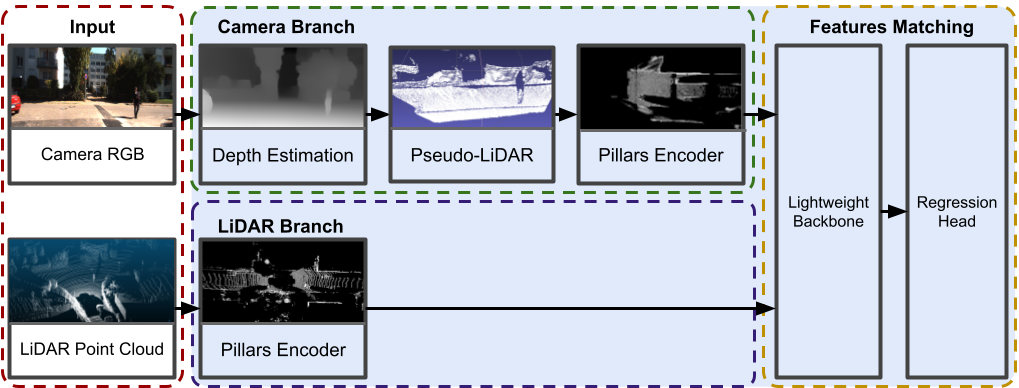}
\end{center}
    \caption{Illustration of the PseudoPillars module, key component of the PseudoCal method allowing for calibration estimation quasi-independently from initial values.}
    \label{fig:pseudoPillars}
\end{figure}

\begin{figure}[t]
  \begin{center}
    \subfloat[Exemple of a pitch decalibration $>20^\circ$]{
       \includegraphics[width=0.49\textwidth]{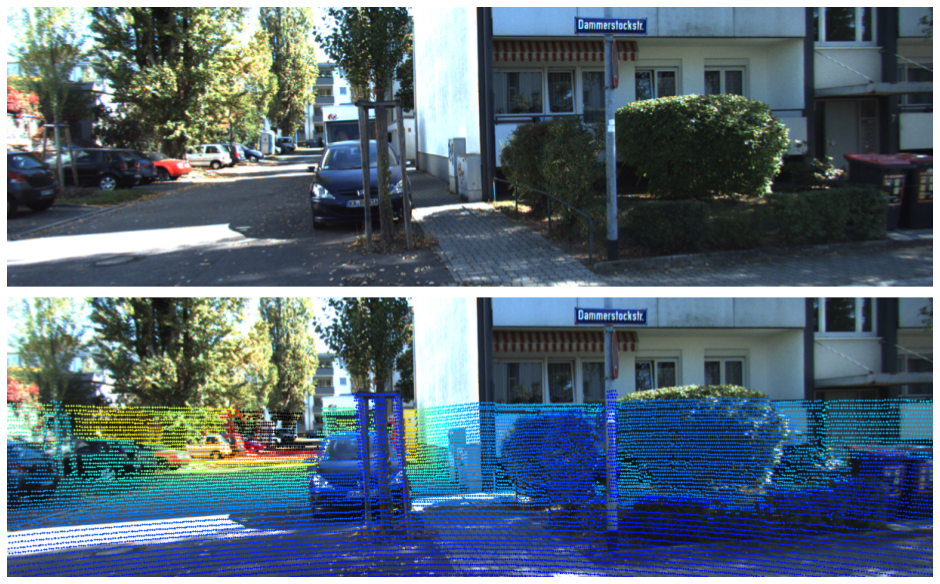}\label{fig:pitch_decal}}
       \hfill 
   \subfloat[Exemple of a yaw decalibration $\approx90^\circ$]{
       \includegraphics[width=0.49\textwidth]{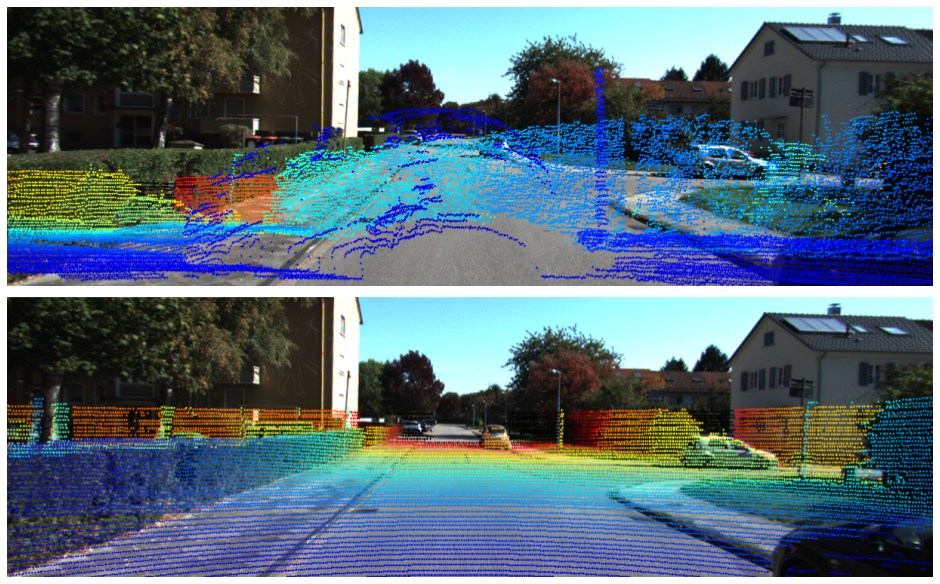}}
\end{center}
\caption{Illustration of the method on driving scenes from KITTI. Points are a depth-coloured LiDAR projection. On top are decalibrated samples in two extreme scenarios where existing deep learning-based methods fail. Bottom visuals represent PseudoCal's correction. (a) shows no LiDAR-camera overlap due to severe pitch axis decalibration, and (b) presents a severe yaw axis decalibration which results in projected points corresponding to objects outside of the camera field of view.}
\label{fig:res}
\end{figure}

In autonomous vehicle and robotics contexts, calibration without an initial estimate is critical. It facilitates on-the-fly recalibration, essential when initial parameters are unavailable or post-mechanical stress like accidents or maintenance, as well as in cases where a robot might be physically unreachable. This adaptability ensures the reliability and precision of the fused sensor data. Furthermore, it accelerates the integration of new sensors, eliminating the need for laborious manual calibration, thus promoting faster adaptation within the autonomous vehicle ecosystem.

The main contribution of this paper lies in a novel camera-to-LiDAR self-calibration technique that effectively leverages the pseudo-LiDAR concept through our proposed PseudoPillars module, as depicted in Figure~\ref{fig:pseudoPillars}. This method is able to perform calibration quasi-independently of initial parameter estimates in typical autonomous vehicle and robotic contexts, addressing extreme cases that remain unsolved by existing approaches (refer to Figure~\ref{fig:res} for examples).

\section{Related Work}

In this section, we provide an overview of calibration methods, focusing on deep learning-based techniques, and highlighting the similarities and differences with our approach. We also discuss the development of monocular depth estimation and pseudo-LiDAR, emphasising their relevance to our work.

\subsection{Automatic Camera-LiDAR Calibration Methods}
Classical computer vision approaches have been used to address this task and have achieved satisfactory accuracy. However, these methods have drawbacks, such as requiring a target~\cite{beltranAutomaticExtrinsicCalibration2022, zhouAutomaticExtrinsicCalibration2018,geigerAutomaticCameraRange2012}, specific environment features~\cite{yuanPixellevelExtrinsicSelf2021}, and for most of them offline and relatively long computation (seconds to minutes)\cite{yuanPixellevelExtrinsicSelf2021,ishikawaLiDARCameraCalibration2018,geigerAutomaticCameraRange2012,beltranAutomaticExtrinsicCalibration2022}.
Deep learning-based calibration methods have emerged as powerful tools for addressing sensor calibration challenges due to their ability to capture complex relationships between sensor modalities while using larger parts of the scene than target-based methods. Examples of such methods include RegNet~\cite{schneiderRegNetMultimodalSensor2017}, CalibNet~\cite{iyerCalibNetGeometricallySupervised2018}, SemAlign~\cite{liuSemAlignAnnotationFreeCameraLiDAR2021}, LCCNet~\cite{lvLCCNetLiDARCamera2021}, DXQ-Net~\cite{jingDXQNetDifferentiableLiDARCamera2022}, and UniCal~\cite{cocheteuxUniCalSingleBranchTransformerBased2023}. Each of these methods has made significant contributions. RegNet~\cite{schneiderRegNetMultimodalSensor2017} was the first work to address this task with a deep learning approach that matches the camera image and projected LiDAR, with parameters refined in a cascaded architecture. DXQ-Net~\cite{jingDXQNetDifferentiableLiDARCamera2022} introduced a differentiable pose estimation module and probabilistic modelling of the task, improving accuracy and generalisation. A more recent approach, UniCal~\cite{cocheteuxUniCalSingleBranchTransformerBased2023}, leverages a Transformer\cite{vaswaniAttentionAllYou2017}-based backbone network to bring attention mechanisms to calibration. It achieves state-of-the-art results with a lighter single-branch architecture. 

However, these methods share a common drawback. They rely on a good initial guess of extrinsic parameters, which may not always be available. Therefore, we propose with PseudoCal a procedure to get an accurate calibration quasi-independently from the initial parameters. Details on the architecture are given in Section~\ref{sec:architecture}.

\subsection{Exploiting a 2D camera for 3D information}
\paragraph{Pseudo-LiDAR}
The pseudo-LiDAR concept has emerged as a key technique for bridging the performance gap between image-based and LiDAR-based 3D object detection. It consists in using a depth map obtained from 2D sensors to generate a point cloud by projecting its points in a 3D space. 
 Wang \etal~\cite{wangPseudoLiDARVisualDepth2020} first demonstrated significant performance improvements on image-based 3D detection by converting stereo-based depth maps into pseudo-LiDAR representations. Weng and Kitani~\cite{wengMonocular3DObject2019} then proposed to generate the depth map from a single camera by monocular depth estimation. Our work capitalises on this approach to generate a point cloud from monocular camera images. 

In network feature extraction, point cloud representation varies. Some, like~\cite{qiPointNetDeepLearning2017,wangPseudoLiDARVisualDepth2020}, use unordered points, while others prefer to voxelise it~\cite{qianEndtoEndPseudoLiDARImageBased2020}. We adopt the approach of Lang~\etal~\cite{langPointPillarsFastEncoders2019}, generating Pillars features from voxels, depicted as a 2D pseudo-image. This method is memory efficient, computationally effective, and suited for sparse or semi-dense point clouds, such as those from LiDARs or pseudo-LiDARs.

\begin{figure}[t]
    \centering
    \includegraphics[width=0.85\textwidth]{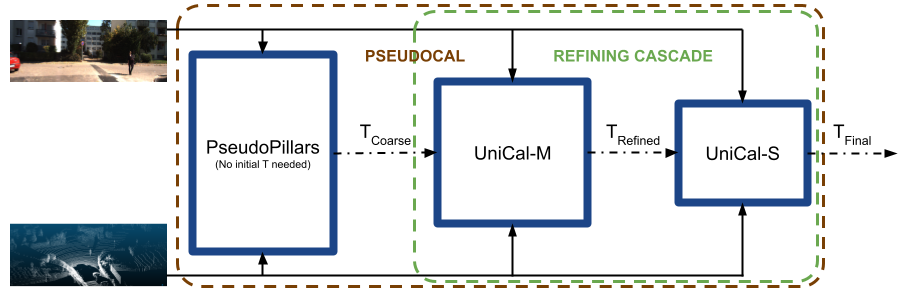}
    \caption{Overview of the PseudoCal architecture and its three main components: PseudoPillars, UniCal-M, and UniCal-S. PseudoPillars provides an initial estimation to the cascaded UniCal modules for refinement. The 6D transformation (calibration parameters) is noted T.}
    \label{fig:pseudocal_arch}
\end{figure}

\paragraph{Monocular Depth Estimation}
Monocular depth estimation is a broadly researched computer vision task. State-of-the-art methods such as \cite{ranftlVisionTransformersDense2021,kimGlobalLocalPathNetworks2022,miangolehBoostingMonocularDepth2021,godardUnsupervisedMonocularDepth2016} now demonstrate reliable and accurate results. These approaches take advantage of advanced techniques and architectures, such as Vision Transformers \cite{ranftlVisionTransformersDense2021}, to capture rich contextual information and model complex scene structures effectively. Moreover, some methods, such as the one proposed by Godard \etal \cite{godardUnsupervisedMonocularDepth2016}, explore unsupervised learning strategies, reducing the need for large-scale annotated depth datasets and further broadening the applicability of monocular depth estimation. These methods have become more reliable and can now be used as critical components for complex vision pipelines. We use a recent model, Global-Local Path Networks (GPLN)~\cite{kimGlobalLocalPathNetworks2022}, as a base component of our PseudoPillars module to provide depth estimation for generating a pseudo-LiDAR projection. GPLN relies on a Transformer-based architecture to capture the global context of the image while using a novel decoder to consider local connectivity.

\section{Method}
In this section, we present the methodology of the proposed PseudoCal approach, which consists of our novel PseudoPillars network, followed by a cascade of two UniCal~\cite{cocheteuxUniCalSingleBranchTransformerBased2023} networks.

\subsection{Architecture}
\label{sec:architecture}

\subsubsection{Cascaded Architecture Rationale}
\label{sec:cascade_rationale}

The PseudoCal architecture adopts a cascaded structure illustrated in Figure~\ref{fig:pseudocal_arch}, comprising the proposed PseudoPillars module followed by two UniCal~\cite{cocheteuxUniCalSingleBranchTransformerBased2023} modules trained on decreasing decalibration ranges as described in Table~\ref{tab:ranges}.The PseudoPillars module performs a coarse estimation of the calibration, while the UniCal modules sequentially refine the estimation. This design choice is motivated by the success of cascaded architectures in various computer vision tasks, including object detection~\cite{caiCascadeRCNNDelving2017}, pose estimation~\cite{chengCascadedParallelFiltering2019}, and even camera to LiDAR calibration~\cite{schneiderRegNetMultimodalSensor2017}. The cascading approach enables PseudoCal to achieve state-of-the-art calibration results quasi-independently from initial parameters.

\begin{table}[t]
    \begin{center}
    \begin{tabular}{ccc}
    \hline 
        \multirow{2}{*}{Module} & \multicolumn{2}{c}{Training decalibration range} \\ 
        & Rotation ($^{\circ}$)                & Translation (cm) \\\hline
         PseudoPillars & 30, 30, 180 & 150\\
         UniCal-M & 10, 10, 10 & 100\\
         UniCal-S & 1, 1, 1 & 10\\
    \hline
    \end{tabular}
    \end{center}
    \caption{Training decalibration range for each cascaded module in PseudoCal. Translation range is the same on all axes, while rotation is respectively for roll, pitch, and yaw axes.}
    \label{tab:ranges}
\end{table}

\subsubsection{The PseudoPillars module}

\begin{figure}[t]
    \centering    \includegraphics[width=0.7\columnwidth]{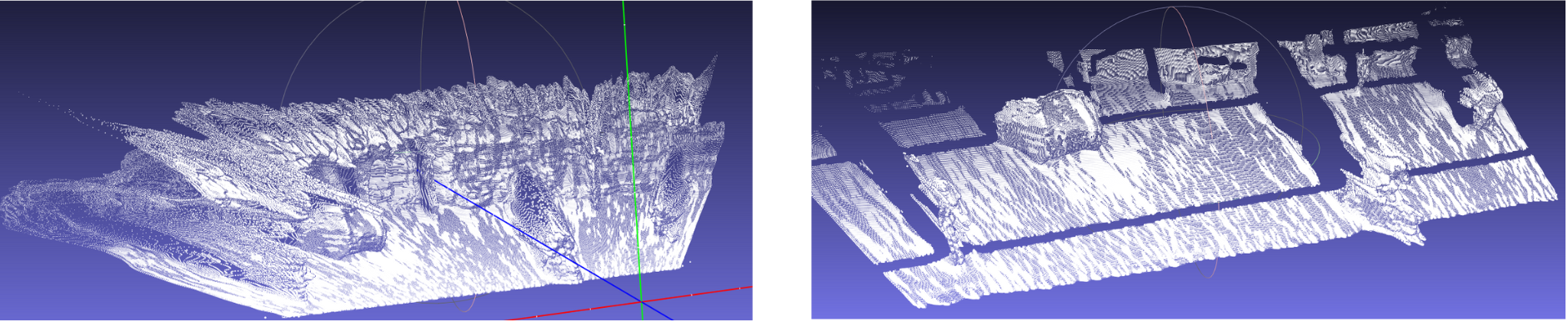}
    \caption{Kitti scene without (left) and with (right) edge removal.}
    \label{fig:canny}
\end{figure}

The PseudoPillars module described in Figure~\ref{fig:pseudoPillars} is the core component of our PseudoCal architecture. It comprises a depth estimation network, a pseudo-LiDAR projection, a Pillars~\cite{langPointPillarsFastEncoders2019} encoder, a matching lightweight backbone (here MobileViT), and a regression head similar to the one used in~\cite{cocheteuxUniCalSingleBranchTransformerBased2023}.

The depth estimation network is based on GLPN~\cite{kimGlobalLocalPathNetworks2022}, which has shown state-of-the-art performance in monocular depth estimation. The estimated depth map is then converted into a pseudo-LiDAR point cloud. This is done by projecting the depth-encoded pixels from the depth map to the 3D space, considering the camera intrinsics. The 3D points coordinates $(x,y,z)$ are given by:
\begin{equation}
\quad z=D(u, v), 
\quad x=\frac{\left(u-c_U\right) \times z}{f_U}, 
\quad y=\frac{\left(v-c_V\right) \times z}{f_V}
\end{equation} where D is the depth map, $\left(u, v\right)$ the pixel coordinates, $\left(c_U, c_V\right)$ the camera center, and$\left(f_U, f_V\right)$ the focal length along U and V axes respectively.

\begin{figure}[t]
     \begin{center}
     \subfloat[Calibrated]{
         \includegraphics[width=0.23\textwidth]{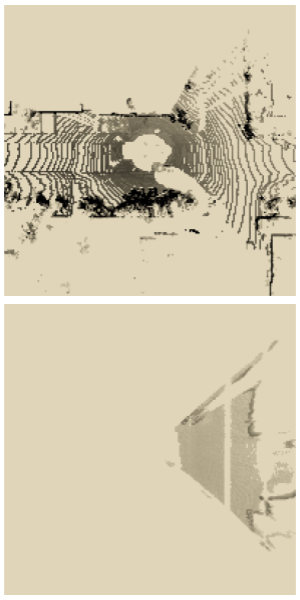}} \hspace{5mm}
     \subfloat[Decalibrated]{ \label{fig:decal}
         \includegraphics[width=0.23\textwidth]{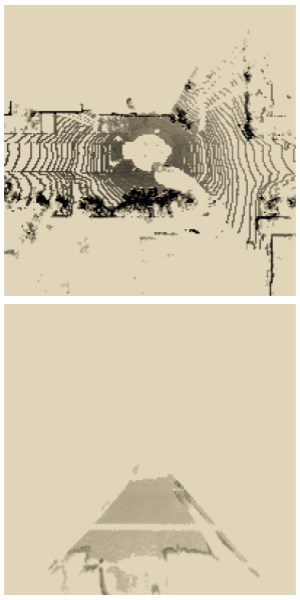}}  \hspace{5mm}
     \subfloat[Decalibrated and augmented]{ \label{fig:augmented}
         \includegraphics[width=0.23\textwidth]{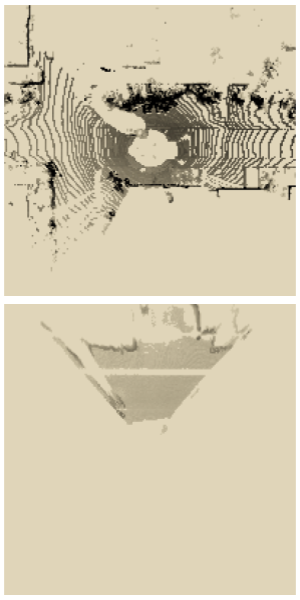}}
         \end{center}

    \caption{Visualization of the Pillars representation of a KITTI scene for the LiDAR (top) and the camera (bottom). (a) is calibrated, (b) is decalibrated by $90^\circ$ on yaw for the illustration, (c) has the same decalibration and a $180^\circ$ yaw augmentation.}
    \label{fig:Pillars}
\end{figure}

This 3D back-projection results in unwanted artefacts on objects' edges, affecting the output of our network. More specifically, trails of points making the junction between 3D objects on different depth, which have already been observed in~\cite{wangPseudoLiDARVisualDepth2020}. Noticing that they are caused by gradient irregularities (inherent to its 2D representation) on the edges in the depth map, we found a simple yet efficient way to filter them. We use the Canny edge detection~\cite{cannyComputationalApproachEdge1986} algorithm to find and remove edges in the depth map before the 3D back-projection, resulting in a clean point cloud, as illustrated in Figure~\ref{fig:canny}. Other methods such as statistical 3D neighbourhood filtering have been considered but have produced lesser results. Other filters could be considered in future research.
To efficiently match information between pseudo-LiDAR and LiDAR, a representation robust to point cloud density is needed. 
Thus, both pseudo-LiDAR and LiDAR point clouds are passed through Pillars encoders, a technique inspired by the Pillar Feature Network used in LiDAR-based 3D object detection~\cite{langPointPillarsFastEncoders2019}. The Pillars encoder generates a compact and efficient feature representation of the point cloud, of which sampling and density can be tuned. Figure~\ref{fig:Pillars} illustrates the Pillars representation obtained from both modalities and highlights their common information.

Both are then jointly passed through a MobileViT~\cite{mehtaMobileViTLightweightGeneralpurpose2022} backbone, which is responsible for extracting useful features for the regression head. MobileViT is a variant of the Vision Transformer~\cite{dosovitskiyImageWorth16x162021} architecture. It has shown impressive performance in various computer vision tasks~\cite{mehtaMobileViTLightweightGeneralpurpose2022}, including camera to LiDAR calibration~\cite{cocheteuxUniCalSingleBranchTransformerBased2023}. 
It has the advantage of being lightweight, fast, and to leverage Transformers and convolutional operations, which allows it to perceive global information in the image, while also considering local features. The regression head then estimates the extrinsic parameters. It is composed of a common dense layer which then forks to two other dense layers to output rotation and translation parameters.

\subsubsection{Refinement cascade}

Following the PseudoPillars module, we cascade two UniCal~\cite{cocheteuxUniCalSingleBranchTransformerBased2023} modules to refine the calibration estimation, as illustrated in Figure~\ref{fig:pseudocal_arch}. UniCal is a model which calibrates a camera-LiDAR pair given an approximate initial calibration, specialized on correcting small decalibrations. Unlike other methods, it fuses camera and LiDAR data early in the process, aggregating image
channels and LiDAR mappings into a unified representation
for joint feature extraction. This approach results in state-of-the-art performance while offering a lightweight solution ideal for resource-constrained applications. 
In PseudoCal, the UniCal modules are trained on two different ranges of decalibrations on which they will specialize. These are described in Table~\ref{tab:ranges} and noted respectively UniCal-M for the medium range ($\pm10^\circ$ and $\pm100cm$), and UniCal-S for the smaller range ($\pm1^\circ$ and $\pm10cm$). As motivated in Section~\ref{sec:cascade_rationale}, by organising the UniCal modules in a cascade, we can refine the coarse output from PseudoPillars to achieve state-of-the-art results reported in Table~\ref{tab:sota} while being quasi-independent from initial parameters. Ablation studies presented in Section~\ref{sec:ablation} and Table~\ref{tab:ablation} support the choice of using exactly two refining UniCal modules.

\subsection{Training Strategy}
\subsubsection{Loss Function}
Our training strategy involves a combination of translation, rotation, and spatial losses to effectively train the PseudoCal network. Translation and rotation losses $\mathcal{L}_t$ and $\mathcal{L}_r$ ensure that the estimated extrinsic parameters are accurate, while spatial losses, inspired by NetCalib~\cite{wuThisWaySensors2021} and CalibNet~\cite{iyerCalibNetGeometricallySupervised2018}, take into account geometric information to achieve better learning convergence.

For translation and rotation losses, we use the mean squared error (MSE) between the predicted and ground-truth values. These two losses are balanced with appropriate weighting factors to account for the difference in their magnitudes. The first spatial loss, $\mathcal{L}_{pcl}$, measures the average distance between corresponding points in the predicted and ground-truth point clouds, which is correlated to the rotation error. The second spatial loss, $\mathcal{L}_C$, is the distance between the centroids of the predicted and ground-truth point clouds, which is correlated with the translation error.
The global loss is defined in Equation~\ref{equation:loss} where weights are defined as $\alpha=1.3,~\beta=1.3,~\gamma=1,~\delta=1.75$.
\begin{equation}
\label{equation:loss}
\mathcal{L} = \alpha \times \mathcal{L}_t + \beta \times\mathcal{L}_r + \gamma \times\mathcal{L}_{pcl} + \delta \times\mathcal{L}_C
\end{equation}

\subsubsection{Samples Generation}

During training, two complementary processes are applied to prevent the model from learning biases. First, artificial decalibration changes the
initial calibration parameters (relative 3D transformation).
It is used to generate different samples to train the network
(within the specified ranges). It helps to generalize, as the number of real setups available is limited
(KITTI setup in our case). On the other side, augmentation
is an absolute 3D transformation applied on both sensors
(no change of target calibration parameters). Augmentation is needed with PseudoPillars to generalise to any orientation of the sensors in its internal 3D space (Pillars representation).

\paragraph{Artificial Decalibration}
To train the model, we need a large number of ground-truth values. It is impossible to get enough values naturally, since one value corresponds to a camera-LiDAR pair; thus we have to adopt an approach similar to~\cite{schneiderRegNetMultimodalSensor2017,cocheteuxUniCalSingleBranchTransformerBased2023} to generate artificial decalibrations on rotation and translation parameters, which we illustrate in Figure~\ref{fig:decal}. 
As we want our network to be able to calibrate a LiDAR and a camera whatever their location on the vehicle or robot, we need to choose decalibration ranges capable of simulating virtually any pairing on the vehicle. Considering a car equipped with a rotating lidar and a camera facing any direction around the car (e.g. front, back, side, etc.), we choose a decalibration of $\pm180^\circ$ on the yaw axis, $\pm30^\circ$ on the other rotation axes, and $\pm150cm$ along each translation axis. Within this setup, the whole yaw rotation is covered. The decalibration range on the other rotation axes is more than any previous works, such as \cite{schneiderRegNetMultimodalSensor2017,cocheteuxUniCalSingleBranchTransformerBased2023,iyerCalibNetGeometricallySupervised2018,wuThisWaySensors2021,jingDXQNetDifferentiableLiDARCamera2022,lvLCCNetLiDARCamera2021,liuSemAlignAnnotationFreeCameraLiDAR2021}. It is amply sufficient to cover most situations : with a $30^\circ$ rotation on the pitch axis, there is not any overlap left between Lidar and camera field of view, as shown in Figure~\ref{fig:pitch_decal}. Similarly, $150cm$ for each translation axis is enough to cover most cases, with, for example, the maximal distance between the LiDAR and a camera in KITTI being $60cm$.

\paragraph{Data Augmentation}
Data augmentation is applied to further improve the generalisation capabilities of PseudoCal. Specifically, we apply a same rotation and translation on point clouds from both sensors, which is illustrated in Figure~\ref{fig:augmented}. This does not affect the calibration parameters (the transformation $T$ between these two sensors), but alters the input for the backbone. It helps ensure that the model does not overfit to a specific sensor configuration or a particular pattern of decalibration.

\paragraph{Training Details}
The loss weights, learning rate ($3e^{-5}$), and batch size (8) were chosen empirically by doing a sweep across a set of values. The different modules were trained independently from scratch and the weights frozen. Training was done on Nvidia V100 GPUs.

\section{Experiments}
In this section, we present a comprehensive set of experiments which primary goals are threefold: (i) to validate the performance of PseudoCal in terms of calibration accuracy, (ii) to evaluate the design choices and the contribution of each module, and (iii) to compare PseudoCal's performance with existing state-of-the-art methods.

\subsection{Dataset}
To evaluate the performance of PseudoCal and compare it to existing works, we employ the KITTI dataset \cite{geigerVisionMeetsRobotics2013}. It provides accurate ground-truth values for extrinsic calibration parameters, making it a reference benchmark for this task. The KITTI dataset comprises real-world data collected by a vehicle equipped with a Velodyne HDL-64E LiDAR sensor and front cameras. We use the same split for training and testing data as the one used in~\cite{schneiderRegNetMultimodalSensor2017}.

\subsection{Results}
\label{sec:results}

\subsubsection{Qualitative Results}
\begin{figure}[t]
    \centering
    \includegraphics[width=1\textwidth]{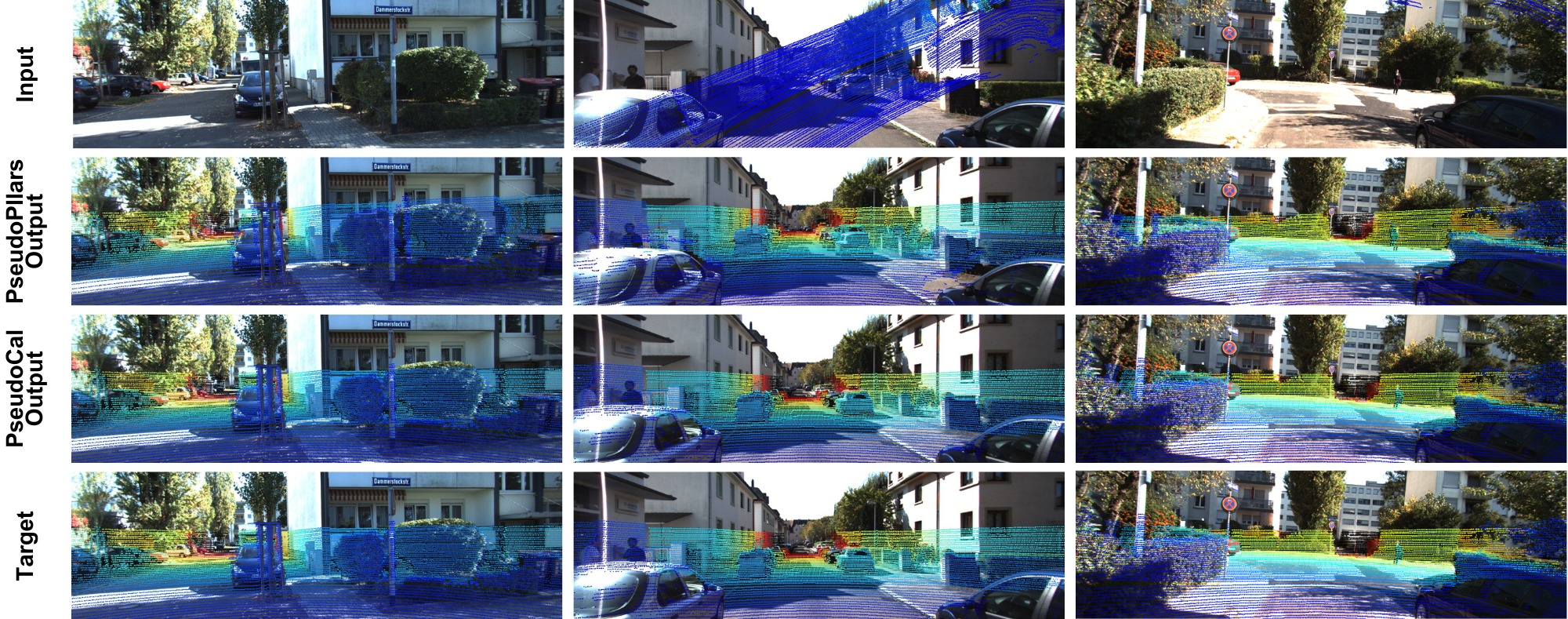}
    \caption{Qualitative evaluation of PseudoCal on KITTI (best viewed on screen). The point clouds are color-coded according to depth. Rows represent the initial decalibrations, PseudoPillars' coarse adjustments, PseudoCal's final refined calibrations, and groundtruth.}
    \label{fig:qualit}
\end{figure}
Figure~\ref{fig:qualit} illustrates the effectiveness of PseudoCal at different stages of the calibration process. The first row shows the initial decalibration, while the second row presents a marked improvement achieved by the PseudoPillars module. This sets the stage for the final row, where the refinement cascade fine-tunes the calibration to a level visually indistinguishable from the ground truth. This demonstrates the efficacy of our two-step approach—coarse adjustment followed by fine-tuning—in achieving accurate calibration, even in extreme cases.

\subsubsection{Comparison to the State of the Art}

\begin{table}[t]
\begin{center}
\begin{tabular}{ccccc}
\hline
\multirow{2}{*}{Model} & \multicolumn{2}{c}{Mean Absolute Error}  & \multicolumn{2}{c}{Decalibration Range} \\
                       & Rotation ($^{\circ}$)                & Translation (cm)                  & Rotation ($^{\circ}$)    & Translation (cm) \\ \hline
RegNet~\cite{schneiderRegNetMultimodalSensor2017}                 & 0.28                     & 6                           & 20, 20, 20  & 150         \\   
CalibNet~\cite{iyerCalibNetGeometricallySupervised2018}               & 0.41                    &    4.34                    & 10, 10, 10  & 20          \\
LCCNet~\cite{lvLCCNetLiDARCamera2021}                 & \textbf{0.03}                    & \textbf{0.36}                     & 20, 20, 20  & 150         \\
DXQ-Net~\cite{jingDXQNetDifferentiableLiDARCamera2022}                & 0.04            & 0.77                   & 5, 5, 5     & 10          \\
UniCal~\cite{cocheteuxUniCalSingleBranchTransformerBased2023}                 & 0.04             & 0.89                  & 1, 1, 1     & 10          \\
PseudoCal (ours)       & 0.05             & 1.18                    & \textbf{30, 30, 180} & \textbf{150}        \\
\hline
\end{tabular}
\end{center}
\caption{Comparison on Mean Absolute Error (MAE) with deep learning-based methods from the state of the art. Rotation decalibration values correspond to the roll, pitch, and yaw axes. Translation decalibration has the same range on all axes. Evaluations are made on different subsets of KITTI. PseudoCal is evaluated on the same set as RegNet, which~\cite{cocheteuxUniCalSingleBranchTransformerBased2023} demonstrate to be the most challenging.} 
\label{tab:sota}
\end{table}

Table~\ref{tab:sota} presents the quantitative results of the proposed PseudoCal method compared to state-of-the-art deep learning-based self-calibration techniques. We report the Mean Absolute Error (MAE) rotation and translation estimates, as well as the acceptable range of decalibration for each method. From these results, we can observe that PseudoCal achieves competitive performance compared to existing state-of-the-art methods~\cite{cocheteuxUniCalSingleBranchTransformerBased2023,jingDXQNetDifferentiableLiDARCamera2022,lvLCCNetLiDARCamera2021,iyerCalibNetGeometricallySupervised2018,schneiderRegNetMultimodalSensor2017}, while being able to deal with the strongest decalibration ranges of all.
The unmatched range of decalibration (up to 180 degrees on the yaw axis) used in our experiments highlights PseudoCal's ability to perform calibration for any camera location without initial information, within usual robotics and autonomous vehicles contexts. Thus, it succeeds in extreme cases where the other compared methods would inherently fail, as illustrated in Figure~\ref{fig:res}.

In summary, PseudoCal is to our knowledge the first deep-learning based extrinsic calibration method that does not focus only on parameters refinement. This makes PseudoCal a step forward for camera-LiDAR calibration in autonomous systems.

\subsubsection{Ablation Study}
\label{sec:ablation}

\begin{table}[t]
\begin{center}
\begin{tabular}{llcc}
\hline
\multicolumn{2}{l}{\multirow{2}{*}{Experiment}}                                                   & \multicolumn{2}{c}{Mean Average Error} \\
\multicolumn{2}{c}{}                                                                              & Rotation ($^{\circ}$)              & Translation (cm)              \\ \hline 

1 & PseudoPillars                                                                            & 3.09              & 19.9              \\ 
2 & \begin{tabular}[t]{@{}l@{}}PseudoPillars without Canny-based Noise Removal\end{tabular}                  & 4.18            & 27.93                \\ 
3 & \begin{tabular}[t]{@{}l@{}}PseudoPillars + UniCal-M\end{tabular}                  & 0.90            & 1.37                \\ 

4 & \begin{tabular}[t]{@{}l@{}}PseudoPillars + UniCal-M + UniCal-$\alpha$\end{tabular} & 0.15                & 1.25              \\ 
5 & \begin{tabular}[t]{@{}l@{}}PseudoPillars + UniCal-M + UniCal-$\alpha$+Unical-S\end{tabular} & 0.05                & 1.19           \\ 
6 & \begin{tabular}[t]{@{}l@{}}\textbf{PseudoPillars + UniCal-M + UniCal-S}\end{tabular} & \textbf{0.05}              & \textbf{1.18}              \\ 
\hline
\end{tabular}
\end{center}
\caption{Ablation experiments results. UniCal-$\alpha$ is trained on a decalibration range of $\pm3^\circ$ for rotation axes and $\pm25cm$ for translation axes.}
\label{tab:ablation}
\end{table}

Results of the conducted experiments are reported in Table~\ref{tab:ablation}.
Experiment 1 evaluates the performance of the PseudoPillars module alone.  It shows, as expected, a higher MAE compared to the complete PseudoCal method. Nevertheless, this is a reliable coarse estimate for the refining cascade.

Experiment 2 shows the efficiency of our Canny-based noise removal (illustrated in Figure~\ref{fig:canny}) in improving accuracy, as not using it leads to an increase of the average error of about $35\%$ on rotation and $40\%$ on translation.

In Experiment 3, we incorporate one UniCal-M module after the PseudoPillars module from Experiment 1. The addition of a single UniCal-M module dramatically improves the calibration performance, with the rotation MAE reduced to $0.90^{\circ}$ and the translation MAE at $1.37cm$. This highlights the effectiveness of the cascading architecture in refining the calibration. This also confirms that the training range for UniCal-M is sufficiently large to accommodate the outputs of PseudoPillars.
In Experiment 4, we tried adding a UniCal-$\alpha$, trained on a decalibration range of $\pm3^\circ$ for rotation axes and $\pm25cm$ for translation axes, which could be a good intermediate range between UniCal-M and UniCal-S. We then added a final UniCal-S in Experiment 5. Finally, we compared it to the actual PseudoCal architecture in Experiment 6, which requires only two refining modules.

Higher MAE in Experiment 4 compared to Experiment 6 suggest that a module trained on a smaller range, such as UniCal-S, is required to achieve state-of-the-art accuracy. Moreover, similar MAE in Experiments 5 and 6 demonstrate that UniCal-M and UniCal-S are sufficient to correct all samples, as an additional intermediate network does not improve the final accuracy. These results thus demonstrate Experiment 6 as the most suitable architecture.

\section{Conclusion}
We have introduced PseudoCal, a novel sensor calibration method that exploits the potential of pseudo-LiDAR through the PseudoPillars module, coupled with a cascaded architecture. This technique enables accurate calibration quasi-independently from any initial knowledge of extrinsic parameters, representing a significant breakthrough in camera-LiDAR calibration. PseudoCal has proven its efficacy on the KITTI dataset, demonstrating its robustness in autonomous system applications. Its architecture limits the number of cascaded modules compared to previous methods~\cite{schneiderRegNetMultimodalSensor2017,lvLCCNetLiDARCamera2021}, and rely on a light refining model~\cite{cocheteuxUniCalSingleBranchTransformerBased2023}, making it an appealing choice for embedded applications.

Future work will investigate extending our approach to other sensor modalities, incorporating additional pseudo-LiDAR representations, and further refining the network architecture and training strategies. We could also consider processing successive frames as sequences to improve the results. 
PseudoCal, with its novel approach to sensor calibration, not only advances the field, but also lays a robust foundation for future research.

\section*{Acknowledgements}
This work was granted access to the HPC resources on the supercomputer Jean Zay of IDRIS under the allocation 2023-AD011014065 made by GENCI. \\ \\
This work has been carried out within SIVALab, joint laboratory between Renault and Heudiasyc (CNRS / Université de technologie de Compiègne).

\bibliography{PseudoCal}
\end{document}